\documentclass[10pt,twocolumn,letterpaper]{article}

\usepackage{cvpr}              %

\usepackage{graphicx}
\usepackage{subcaption}
\usepackage{float}
\usepackage{caption}
\usepackage{lscape}                                         %
\usepackage{wrapfig}

\usepackage[british,american]{babel}

\usepackage[lined,ruled,linesnumbered]{algorithm2e}
\usepackage{animate} %

\usepackage{booktabs}                   %
\usepackage{multirow}

\usepackage{enumitem}

\usepackage{bm}                          %

\usepackage{amssymb}
\usepackage{amsmath}  %

\usepackage{units}
\usepackage{color}
\usepackage[T1]{fontenc}    %
\usepackage{amsfonts}       %
\usepackage[utf8]{inputenc} %
\usepackage{nicefrac}       %
\usepackage{microtype}      %
\usepackage{pifont}         %
\usepackage{comment}

\usepackage[mathscr]{euscript}
\usepackage{colortbl}

\usepackage{url}  %

\usepackage[pagebackref=true, breaklinks=true,letterpaper=true,colorlinks,bookmarks=false]{hyperref}

\def\vs{\emph{vs.}~}                 %

\newcommand{\mfigure}[2]
{
\includegraphics[width=#1\linewidth]{#2}
}

\newcommand{\secref}[1]{Section~\ref{sec:#1}}
\newcommand{\figref}[1]{Figure~\ref{fig:#1}} 
\newcommand{\tabref}[1]{Table~\ref{tab:#1}}

\long\def\ignorethis#1{}

\newcommand {\torevise}[1]{#1}

\newcommand {\added}[1]{{#1}}

\newbox\jsavebox%

\def\img{\bm{x}}
\def\gt{\bar{\bm{y}}}
\def\out{{\bm{y}}}

\def\feat{\bm{f}}
\def\hidden{\bm{h}}
\def\loss{{L}}

\def\encs{{E}}
\def\decs{{D}}

\graphicspath{{figure}, {example}}

\usepackage{xcolor}
\colorlet{dark-blue}{blue!50!black}
\colorlet{dark-cyan}{cyan!75!black}
\colorlet{dark-purple}{purple!50!black}
\colorlet{dark-red}{red!75!black}
\colorlet{dark-green}{green!75!black}
\colorlet{dark-orange}{orange!50!black}
\colorlet{dark-gray}{black!75}
\colorlet{light-gray}{black!30}
\definecolor{nice-red}{HTML}{E41A1C}
\definecolor{nice-orange}{HTML}{FF7F00}
\definecolor{nice-yellow}{HTML}{FFC020}
\definecolor{nice-green}{HTML}{39b54a}
\definecolor{nice-blue}{HTML}{0071bc}
\definecolor{nice-purple}{HTML}{984EA3}

\colorlet{verylight-gray}{black!10}
\definecolor{LightCyan}{rgb}{0.66,0.85,0.76}

\makeatletter
\def\@fnsymbol#1{\ensuremath{\ifcase#1\or \dagger\or \ddagger\or
\mathsection\or \mathparagraph\or \|\or **\or \dagger\dagger
\or \ddagger\ddagger \else\@ctrerr\fi}}
\makeatother

\graphicspath{{figures/}}

\hypersetup{
    citecolor=dark-green,%
    filecolor=blue,
    linkcolor=nice-red,
    urlcolor=nice-red
}

\usepackage[capitalize]{cleveref}
\crefname{section}{Sec.}{Secs.}
\Crefname{section}{Section}{Sections}
\Crefname{table}{Table}{Tables}
\crefname{table}{Tab.}{Tabs.}

\def\cvprPaperID{2614} %
\def\confName{CVPR}
\def\confYear{2022}

\begin{document}

\title{Neural Global Shutter: Learn to Restore Video\\
from a Rolling Shutter Camera with Global Reset Feature}

 \author{Zhixiang Wang$^{1,2,3}$
    \hspace{0.08in} Xiang Ji$^{1}$
    \hspace{0.08in} Jia-Bin Huang$^{4}$
    \hspace{0.08in} Shin'ichi Satoh$^{3,1}$
    \hspace{0.08in} Xiao Zhou$^{5}$\thanks{Corresponding authors}
    \hspace{0.08in} Yinqiang Zheng$^{1}$\footnotemark[1]
	\vspace{1mm} \\
	\hspace{0.1in} $^{1}$The University of Tokyo
	\hspace{0.1in} $^{2}$RIISE
    \hspace{0.10in} $^{3}$National Institute of Informatics\\
    \hspace{0.10in} $^{4}$University of Maryland College Park
    \hspace{0.10in} $^{5}$Hefei Normal University
	}

\maketitle

\begin{abstract}

Most computer vision systems assume distortion-free images as inputs. The widely used rolling-shutter (RS) image sensors, however, suffer from geometric distortion when the camera and object undergo motion during capture. Extensive researches have been conducted on correcting RS distortions. However, most of the existing work relies heavily on the prior assumptions of scenes or motions. Besides, the motion estimation steps are either oversimplified or computationally inefficient due to the heavy flow warping, limiting their applicability. In this paper, we investigate using rolling shutter with a global reset feature (RSGR) to restore clean global shutter (GS) videos. This feature enables us to turn the rectification problem into a deblur-like one, getting rid of inaccurate and costly explicit motion estimation.
First, we build an optic system that captures paired RSGR/GS videos. Second, we develop a novel algorithm incorporating spatial and temporal designs to correct the spatial-varying RSGR distortion. Third, we demonstrate that existing image-to-image translation algorithms can recover clean GS videos from distorted RSGR inputs, yet our algorithm achieves the best performance with the specific designs. Our rendered results are not only visually appealing but also beneficial to downstream tasks. Compared to the state-of-the-art RS solution, our RSGR solution is superior in both effectiveness and efficiency. Considering it is easy to realize without changing the hardware, we believe our RSGR solution can potentially replace the RS solution in taking distortion-free videos with low noise and low budget.
\end{abstract}

\section{Introduction}
\label{sec:intro}

\begin{figure*}[!tb]
    \centering
    \mfigure{1}{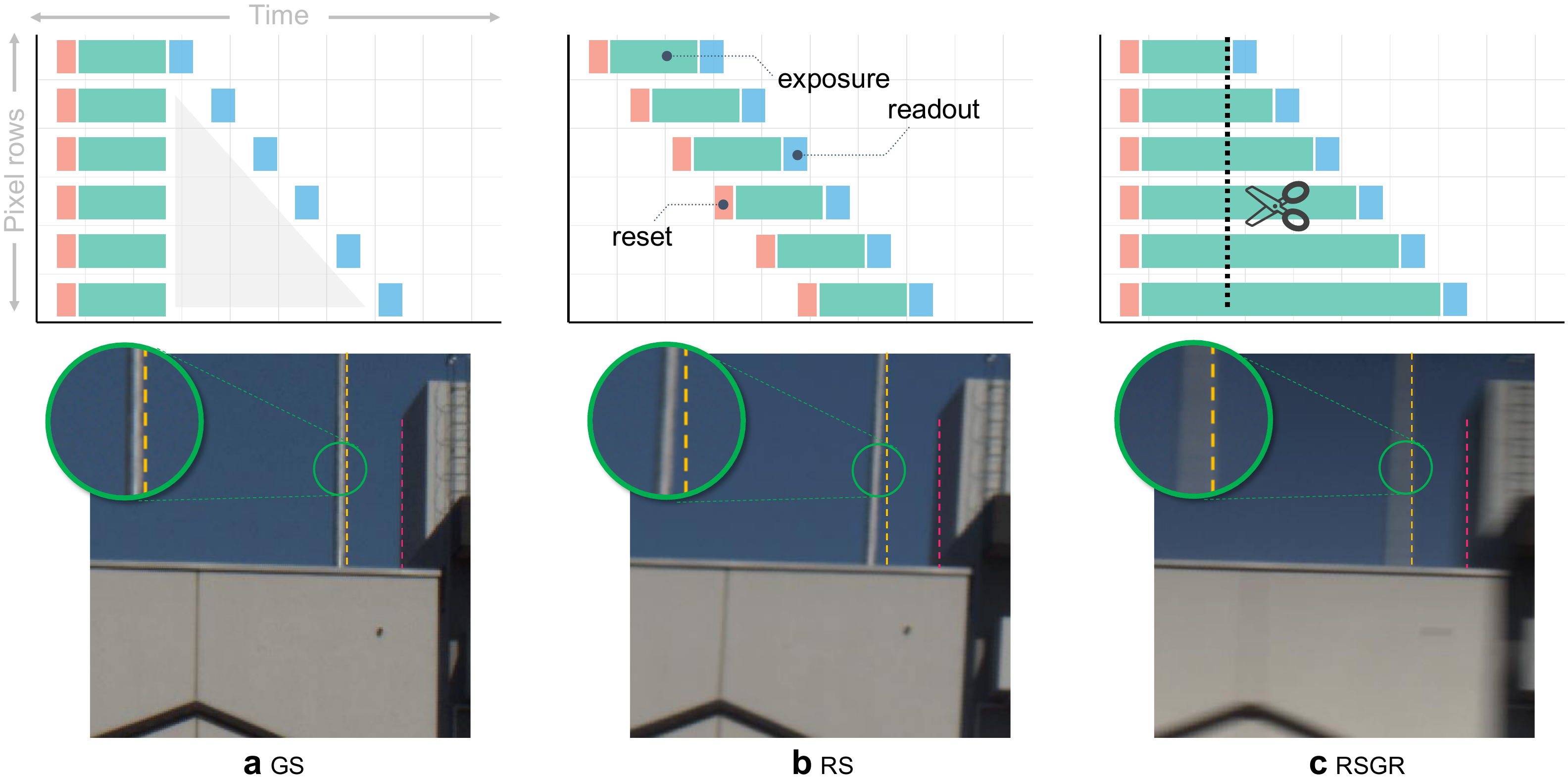}
    \caption{\textbf{Different exposure modes.} \textbf{a}, GS exposes all pixels simultaneously. This manner requires additional memory nodes to store charged pixels before readout so that it is costly and often suffers from noises.
    \textbf{b}, RS exposes pixels scanline by scanline with a time delay, featured in less noise, higher frame rate, and lower cost. Unfortunately, it leads to distortions when the scene or the camera undergoes motion. Reducing these distortions needs to estimate the motion and complement pixels with the estimation. These steps rely on various assumptions and are time-consuming, limiting RS sensors' applicability.
    \textbf{c}, RSGR is a widely ignored feature of RS. It begins exposure of all pixels at the same time and ends them scanline by scanline.
    The varying exposure duration of different scanlines yields spatial-varying brightness and blur.
    RSGR enables us to turn the old RS rectification problem into a deblur-like one.
    }
    \label{fig:problem}
\end{figure*}

Image sensor is the important component converting photons into digital signals, for machine to see \cite{Saunders-Nature-2019-Computational,Aittala-NeurIPS-2019-Computational}, to understand \cite{He-ICCV-2017-mask,Zellers-CVPR-2018-motifs}, and to recreate \cite{Karras-CVPR-2019-style,Mildenhall-ECCV-2020-nerf} the visual world.
It comprises millions of spatially distributed photodiodes, \added{namely \emph{pixels},} performing photoelectric conversion and charge accumulation \added{when photons arrive during exposure duration}.
\added{
Readout circuits read these accumulations out and converting them into spatially distributed digital signal, \ie, image, when pixels complete charging.
Since simultaneously reading all pixels out requires millions of circuits, leading to unaffordable costs, a key design of image sensors is scheduling the exposure time and readout time of different pixels to reuse limited readout circuits.
This function is based on the on-chip electronic shutters dominated by two modes: \emph{global} shutter (GS) and \emph{rolling} shutter (RS).
}

Image sensors with different on-chip electronic shutters hold contrast characteristics.
\added{GS-based image sensors expose all pixels simultaneously and transfer accumulated charges to a storage area
before readout. In this way, they can read out charges sequentially with a few readout circuits (\figref{problem}\textcolor{nice-red}{a}).
But the requirement for additional storage increases their expense and power consumption and leads to more noises.
Differently, RS-based image sensors expose pixels scanline by scanline with a time delay (\figref{problem}\textcolor{nice-red}{b}). This delay enables RS sensors to overlap exposure and readout time, leading to a higher frame rate. Besides, exemption from
additional storage area attributes to RS sensors lower cost and fewer noises. Unfortunately, they bring
distortions when scenes or cameras undergo motion.}
The faster the movement, the larger the distortion. The distortion obstructs vision systems equipped with RS sensors to precision-sensitive applications, \eg,
localization~\cite{Zamir-ECCV-2010-Localization}, optical flow~\cite{Teed-ECCV-2020-raft}, and reconstruction~\cite{Lensch-TOG-2003-recons}.
It naturally poses a research question: 
{if there exists a solution for taking distortion-\emph{free} videos with \emph{low} noise and \emph{low} budget.}

We categorize existing solutions into two folds: {hardware-based} and {computational}.
Hardware-based solutions implement the GS function on RS sensors by placing additional memory nodes to store pixels on the charge \cite{Velichko-2015-CMOS}, the voltage \cite{Stark-2018-Back}, even the digital domain \cite{Seo-2021-VLSI}.
The disadvantages in sensor size, cost and noise are obvious.
Computational methods operate directly on RS outputs by correcting the RS distortions \cite{Liang-TIP-2008,Rengarajan-CVPR-2016,Zhuang-ICCV-2017,Purkait-WACV-2018,Lao-2018-CVPR,Zhuang-CVPR-2019,Liu-CVPR-2020-unrolling,Zhong-arXiv-2022}. 
Though existing methods are different in either the input forms: {single-image} \vs {multi-image}, 
or the method types: {classical} \vs {learning-based},
their basic ideas that complement the distorted pixels through estimating pixel-wise motions are the same.
In the case of \emph{single} image input, due to the ill-posedness, 
classical methods estimate motion based on additional assumptions on the scenes \cite{Rengarajan-CVPR-2016,Lao-2018-CVPR} or the camera motions \cite{Purkait-WACV-2018}.
Learning-based approaches relax the assumptions on scenes by digesting the implicit prior, but those on camera motions \cite{Rengarajan-CVPR-2017,Zhuang-CVPR-2019} still hold.
The assumptions on either scenes or camera motions compromise their practical applicability. When the input comprises \emph{multiple} images, this problem becomes well-posed, since one can directly estimate the motion between two consecutive frames with classical  \cite{Zhuang-ICCV-2017,Zhuang-ECCV-2020} or learning-based methods \cite{Liu-CVPR-2020-unrolling}.
Then, they approximate the displacement between the RS frame and the \emph{virtual} GS frame and use the approximation for warping.
But, they usually cannot work under large and complicated camera motions since complementing information is essentially hard, especially when they encounter \emph{oversimplified} motion models.
Moreover, the motion estimation steps are often computationally \emph{inefficient} and is unacceptable by real-time applications.

\begin{figure*}[!tb]
    \centering
    \mfigure{1}{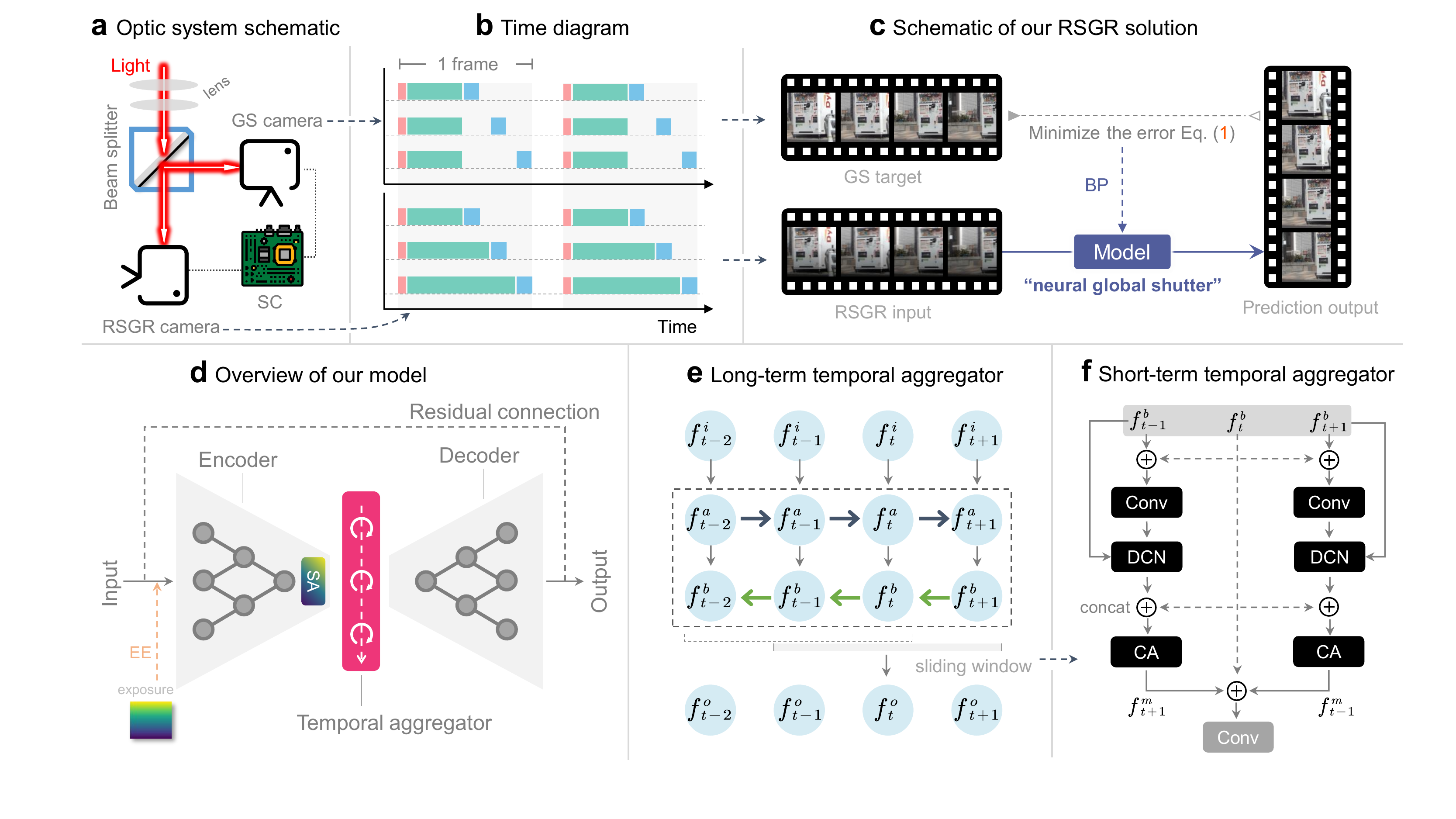}
    \caption{\textbf{The proposed method.} 
    {\textbf{a}, We divide an incoming light into two same parts through a beam splitter and feed these splits into a GS camera and an RSGR camera, respectively. Since the cameras are spatially aligned using a calibration tool and synchronized through synchronization circuits (SC), we can record the light with two different exposure manners.
    \textbf{b}, We keep the first scanline's exposure time of GS and RSGR cameras to be equal.
    \textbf{c}, Given the RSGR video capture, our model outputs a corresponding GS video prediction in an end-to-end fashion.
    We optimize the algorithm using the backpropagation (BP) technique by minimizing the error between the prediction and the target GS video capture.
    As the target GS videos suffer from noises, we carefully chose the loss function to suppress the negative effect.
    \textbf{d}, Our model is based on encoder-decoder structure. We employ EE and SA to generate spatial-sensitive feature maps. Given the processed feature maps, we use dual temporal information aggregators to gather both long-term (\textbf{e}) and short-term temporal information (\textbf{f}). They create powerful feature maps for the decoder to render clean GS videos. To keep the details, we use the residual connection.
    }}
    \label{fig:method}
\end{figure*}

In this paper, we propose a new solution based on a widely ignored feature of RS sensors---\emph{global reset} (GR). This feature allows us to \emph{convert} the old RS rectification problem into a deblur-like one.
Thus, we can throw away inaccurate and time-consuming motion estimation steps and make the problem easier to solve.
It is because RSGR exposes all pixels at the same time like GS rather than scanline by scanline with a constant time delay like conventional RS \cite{Nakamura-2017-Sensor}, as \figref{problem}\textcolor{nice-red}{c} shows.
The varying exposure duration of different scanlines yields spatial-varying blur and brightness when capture undergoes motion.
This distortion is different from pixel shifts of the RS sensor.
Most RS sensors come with this feature that allows them to use mechanical shutters or strobe lights to overcome the distortion, \added{like what Bradley \etal~\cite{Bradley-CVPRW-2009} does to solve the RS distortion}. 
We relax the hardware requirements in a computational way, as shown in \figref{method}.
To facilitate the development and evaluation of data-driven algorithms, we build an optic system to take paired RSGR/GS videos simultaneously. 
This system offers our community a new dataset consisting of 79 paired RSGR/GS video sequences captured under real scenes.
We further propose an new algorithm to tackle the unique distortion.
Our method contains three main components: 
\textbf{1)} a spatial-aware feature encoder that extracts low-dimensional feature representations for each input RSGR frame. It has two special designs: exposure encoding (EE) and spatial attention (SA) ~\cite{Hou-CVPR-2021} for producing spatial-sensitive feature maps. 
\textbf{2)}~a module with two recurrent neural networks (RNNs)~\cite{Rumelhart-1985-RNN} to propagate long-term information along the time axis bidirectionally. 
\textbf{3)}~a module stacked with a few convolutional layers (Conv), a deformable convolutional network (DCN)~\cite{Dai-ICCV-2017-Deformable}, and a channel attention (CA) to fuse neighboring frames. 
We experimentally demonstrate that the widely ignored global reset feature enables us to recover clean GS videos from RS sensors with existing image-to-image translation algorithms.
With specific designs, our algorithm achieves the \emph{best} performance.
Especially, it can work under significant camera motion where correspondences between two consecutive frames are hard to estimate for RS solutions.
Compared to existing RS solutions, our solution without explicit motion estimation is \emph{effective} and \emph{efficient}.
Considering it is easy to realize without changing the hardware, we believe our solution can be an alternative to the RS solution.
To sum up, we make the following three contributions:
\begin{itemize}
\item \textbf{Problem:} we are the {first} to introduce RSGR, a widely ignored feature, to our community. This feature enables us to convert the old RS rectification problem into a deblur-like one.

\item \textbf{Optic system and dataset:} we build an optic system that takes paired RSGR/GS videos and offer a new dataset captured under real scenes.
The large-scale paired dataset enables developing and evaluating data-driven methods.
We release the dataset to facilitate following researches\footnote{\href{https://github.com/lightChaserX/neural-global-shutter}{https://github.com/lightChaserX/neural-global-shutter}}.

\item \textbf{Algorithm:} we propose a {novel} algorithm for RSGR video restoration. 
We experimentally demonstrate that it can render clean GS videos from distorted RSGR inputs and our integrated solution has the potential to replace the RS solution.

\end{itemize}
\section{Method}
\label{sec:method}

\subsection{Paired video acquisition system}

Global reset is a widely ignored feature, and investigating how to restore clean GS videos from RSGR videos is fresh. 
We are not aware of \emph{any} dataset that contains RSGR videos and their corresponding GS counterparts.
As the first attempt to crack this challenging nut, we build an optic system (\figref{method}\textcolor{nice-red}{a}) that captures synchronized RSGR and GS videos to facilitate developing and evaluating new algorithms.
The system employs a beam splitter to divide the incoming light into two parts and feeds them into an RSGR and a GS camera.
These two cameras are spatially calibrated with a calibration tool and synchronized through a synchronization circuit (\figref{method}\textcolor{nice-red}{b}). 
Thus, they can capture the same frames simultaneously. 
Besides, we also ensure the RSGR camera's exposure duration of the first scanline equals the GS camera's exposure duration. This system enables us to develop data-driven algorithms. Note that, due to the installation restrictions in practice, the actual scanning direction of all RSGR frames in this dataset is bottom-to-top. This will not affect the effectiveness of an algorithm trained on thus frames, as will be verified in the generalization evaluation.

\subsection{Neural global shutter}

Thanks to the global reset feature, we recover clean GS videos from RS sensors in a deblur-like way.  
Thus, following the common practice in image/video deblur, we employ the encoder-decoder structure (\figref{method}\textcolor{nice-red}{d}).
Nevertheless, since the challenging RSGR distortion consists of spatial-varying blur and brightness, we make three specific designs, including the spatial-aware encoder, the long-term and short-term temporal information aggregator.

\paragraph{Spatial-aware encoder.}
We use an encoder $\encs_\theta$ to extract the low-dimensional representations $\{\feat_t^i\}_{t=1}^S$ for a given video segment $\{\img_t\}_{t=1}^S$.
Our encoder $\encs_\theta$ operates on each frame respectively.
Two unique designs power its ability to address spatial-varying distortions.
First, we encode each pixel's exposure duration (EE) and feed them along with the input frame into $\encs_\theta$. The input $\img_t$, therefore, has four channels.
This design comes from our observation that the RSGR distortion gradually changes and is relevant to the exposure duration.
Second, we integrate the spatial attention (SA) mechanism \cite{Hou-CVPR-2021} 
into the encoder $\encs_\theta$ for producing spatially selective feature maps, further enabling us to embed the position information.
These two components enable the encoder to be adaptive to the exposure duration.

\paragraph{Long-term temporal aggregator.}
Similar to most of existing video deblur methods use temporal information as an essential cue,
we leverage both the \emph{long-term} and \emph{short-term} temporal information from the input video segment $\{\img_t\}_{t=1}^{S}$.
Specifically, we aggregate long-term temporal information $\{\feat_t^i\}_{t=1}^{S}$ with two RNNs bidirectionally. 
We first perform forward information aggregation, with $t$ increases from 1 to $S$, the output at time $t$ is $\feat_t^a = F_a([\feat_t^i, \hidden_{t-1}^a])$, with $\hidden_{t}^a = H_a(\feat_t^a)$,
where $\hidden_t^a$ is the hidden state, $[\cdot, \cdot]$ denotes concatenation operation along the channel axis\footnote{We set the length of input video segment $S$ to 8.}. We instantiate $F_a$ and $H_a$ with residual blocks (RBs)~\cite{He-CVPR-2016-ResNet} and residual dense blocks (RDBs)~\cite{Zhang-CVPR-2018-RDB}.
The initial hidden state $\hidden_0^a$ is set to 0.
Given the outputs from the forward aggregator, we then perform backward information aggregation, with $t$ decreases from $S$ to 1, the output at time $t$ is $\feat_t^b = F_b([\feat_t^a, \hidden_{t+1}^b])$, with $\hidden_{t}^b = H_b(\feat_t^b)$.
Likewise, $F_b$ and $H_b$ comprise RBs and RDBs.
The initial hidden state $\hidden_{S+1}^b$ is set to 0. We also tried to initialize $\hidden_{S+1}^b$ with $\hidden_{S}^a$, but the results are unsatisfactory. We propagate bidirectionally as we find that using only one-directional temporal information will unbalance different frames. It is even worse than without using long-term temporal information.

\paragraph{Short-term temporal aggregator.}
We have incorporated the long-term temporal information, while the short-term (local) temporal information is also essential.
We use a sliding window to traverse the video segment.
In the window, we implicitly align the center frame feature $\feat_{t}^b$ and its neighboring frame features $\{\feat_{t\pm k}^b\}_{k= 1}^K$ using a deformable convolutional network (DCN)~\cite{Dai-ICCV-2017-Deformable}.
Given the aligned neighboring features and the center feature, we concatenate them along the channel axis. We use the channel attention (CA) to learn to weight them and a convolutional layer (Conv) to learn to fuse them\footnote{We set the length of the window to 3, its stride to 1, and $K$ to 1.}.

\paragraph{Decoder.}
Taking a refined feature map $\feat_t^o$ as input, the decoder $\decs_\theta$ renders a clean GS image $\out_t$. To capture details, we let our network to learn the difference $\Delta \img_t$ between the generated image $\out_t$ and the original image $\img_t$ with a global residual connection \cite{He-CVPR-2016-ResNet}: $\out_t = \img_t + \Delta \img_t$.

\paragraph{Supervision.}
We train our network in a supervised fashion using the target GS videos as the direct supervisory signal.
Specifically, we compute the differences between the rendered video frames $\{\out_t\}_{t=1}^T$ and its GS counterparts $\{\gt_t\}_{t=1}^T$. 
But since GS videos often suffer from noises, we use the perceptual loss~\cite{Johnson-ECCV-2016-Perceptual} and SSIM loss~\cite{Wang-TIP-2004-SSIM} together to suppress the negative effects caused by noisy supervision:
\begin{equation}\label{eq:loss}
    \loss = 
    \lambda\sum_{l}\lVert{\psi_l(\gt_t) - \psi_l(\out_t)}\rVert_1 +  
    (1 - \lambda) \phi(\out_t, \gt_t)\,,
\end{equation}
where $\psi_l$ is the feature extracted from $l$-th layer of a pre-trained VGG-19 network~\cite{Simonyan-2014-VGG};
$\phi(\cdot,\cdot)\in [0, 1]$ is 1 minus a differentiable version of the SSIM~\cite{dSSIM}.
The parameter $\lambda$ controls the balance between perceptual and SSIM components, which is dynamically set.
All of the errors resulting from $T$ frames are summarized together.

\begin{table*}[!tp]
    \centering
    \caption{\textbf{Quantitative comparison}.
    The performance is measured with {mean PSNR/SSIM} (higher is better).
    `F' denotes evaluation using full-size frames, while `U', `M' and `L' represent using only 200 rows of pixel from the top, middle, and bottom of each frame, respectively.
    $^\dagger$Training DSUR on our task from scratch is difficult. Thus we fine-tune it, unlike other algorithms trained from scratch here.
    $^\ddagger$Our model without using temporal information. \textbf{Bold} texts indicate the best method for each metric.
    }
    \resizebox{\linewidth}{!}
    {
    \begin{tabular}{lcccccccc}
        \toprule
        {} & \multicolumn{4}{c}{\textsc{Set-I}} & \multicolumn{4}{c}{\textsc{Set-II}}\\\cmidrule(lr){2-5}\cmidrule(lr){6-9}
        Method & F & U & M & L & F & U & M & L \\\midrule
        Input & 18.95 / 0.75 & 25.32 / 0.82 & 21.56 / 0.81 & 16.36 / 0.63 & 17.82 / 0.73 & 23.64 / 0.77 & 21.45 / 0.77 & 15.54 / 0.66 \\
        deblurGANv2~\cite{Kupyn-ICCV-2019-deblurGANv2}       & 19.97 / 0.73 & 21.54 / 0.75 & 23.73 / 0.77 & 18.17 / 0.69 & 18.34 / 0.69 & 20.14 / 0.69 & 22.14 / 0.71 & 17.28 / 0.66 \\ 
        SRN~\cite{Tao-CVPR-2018-SRN}                  & 26.87 / 0.86 & 26.12 / 0.83 & 27.08 / 0.85 & 29.59 / 0.89 & 25.05 / 0.81 & 24.32 / 0.79 & 25.65 / 0.81 & 27.02 / 0.83 \\
        STRCNN~\cite{Hyun-ICCV-2017-STRCNN}           & 24.88 / 0.85 & 24.27 / 0.83 & 25.33 / 0.85 & 27.54 / 0.88 & 22.59 / 0.81 & 22.99 / 0.79 & 23.46 / 0.81 & 23.66 / 0.83 \\
        DBN~\cite{Su-CVPR-2017-DBN}                   & 26.49 / 0.87 & 26.50 / 0.85 & 26.66 / 0.87 & 28.47 / 0.89 & 22.57 / 0.81 & 23.24 / 0.80 & 23.81 / 0.81 & 23.24 / 0.82 \\
        IFIRNN~\cite{Nah-CVPR-2019-IFIRNN}            & 28.01 / 0.89 & 27.20 / 0.88 & 28.35 / 0.89 & 29.21 / 0.90 & 25.17 / 0.82 & 24.77 / 0.80 & 25.62 / 0.81 & 26.94 / 0.84 \\
        ESTRNN~\cite{Zhong-ECCV-2020} & 25.85 / 0.89 & 26.67 / 0.88 & 30.16 / 0.90 & 25.19 / 0.89 & 22.72 / 0.83 & 23.42 / 0.81 & 26.03 / 0.83 & 22.86 / 0.83\\
        DSUR$^\dagger$~\cite{Liu-CVPR-2020-unrolling} & 24.72 / 0.84 & 24.30 / 0.81 & 25.65 / 0.85 & 26.63 / 0.86 & 22.50 / 0.80 & 22.49 / 0.78 & 23.87 / 0.81 & 23.38 / 0.83 \\
        JCD~\cite{Zhong-CVPR-2021}                    & 28.15 / 0.85 & 27.50 / 0.84 & 28.73 / 0.85 & 30.44 / 0.87 & 25.33 / 0.80 & 24.77 / 0.78 & 25.71 / 0.80 & 27.43 / 0.83 \\\midrule
        Ours-noT$^\ddagger$                           & 27.56 / 0.85 & 26.23 / 0.83 & 27.55 / 0.85 & 31.55 / 0.88 & 25.37 / 0.80 & 24.74 / 0.77 & 25.65 / 0.79 & 27.29 / 0.82 \\
        Ours                                          & \textbf{32.72 / 0.92} & \textbf{31.83 / 0.92} & \textbf{33.01 / 0.92} & \textbf{34.65 / 0.92} & \textbf{27.29 / 0.85} & \textbf{26.96 / 0.84} & \textbf{27.57 / 0.85} & \textbf{28.35 / 0.86}\\
        \bottomrule
    \end{tabular}
    }
    \label{tab:compare-img-sota}
\end{table*}

\newlength{\ablationw}
\setlength{\ablationw}{0.081\textwidth}
\newlength{\ablationh}
\setlength{\ablationh}{0.081\textwidth}
\newcommand{\cubrowy}[2]{            
	\includegraphics[width=2\ablationw,height=2\ablationh]{pngresults/scene#1_#2_input.png} &
    \includegraphics[width=2\ablationw,height=2\ablationh]{pngresults/scene#1_#2_deblurGANunpair.png} &
	\includegraphics[width=2\ablationw,height=2\ablationh]{pngresults/scene#1_#2_srn.png} &
	\includegraphics[width=2\ablationw,height=2\ablationh]{pngresults/scene#1_#2_oursnoT.png} &
	\includegraphics[width=2\ablationw,height=2\ablationh]{pngresults/scene#1_#2_strcnn.png} &
	\includegraphics[width=2\ablationw,height=2\ablationh]{pngresults/scene#1_#2_dbn.png} 
	\\
	Input & 
	deblurGANv2~\cite{Kupyn-ICCV-2019-deblurGANv2} & 
	SRN~\cite{Tao-CVPR-2018-SRN} &
	Ours-noT &
    STRCNN~\cite{Hyun-ICCV-2017-STRCNN} & 
    DBN~\cite{Su-CVPR-2017-DBN} 
    \\
    \includegraphics[width=2\ablationw,height=2\ablationh]{pngresults/scene#1_#2_ifirnn.png} &
    \includegraphics[width=2\ablationw,height=2\ablationh]{pngresults/scene#1_#2_estrnn.png} &
	\includegraphics[width=2\ablationw,height=2\ablationh]{pngresults/scene#1_#2_Unroll.png} &
	\includegraphics[width=2\ablationw,height=2\ablationh]{pngresults/scene#1_#2_jcd.png} &
	\includegraphics[width=2\ablationw,height=2\ablationh]{pngresults/scene#1_#2_ours.png} &
	\includegraphics[width=2\ablationw,height=2\ablationh]{pngresults/scene#1_#2_gt.png}\\
	IFIRNN~\cite{Nah-CVPR-2019-IFIRNN} &
	ESTRNN~\cite{Zhong-ECCV-2020} &
	DSUR~\cite{Liu-CVPR-2020-unrolling} &
	JCD~\cite{Zhong-CVPR-2021} &
    Ours &
    GT
}
\begin{figure*}[h!]
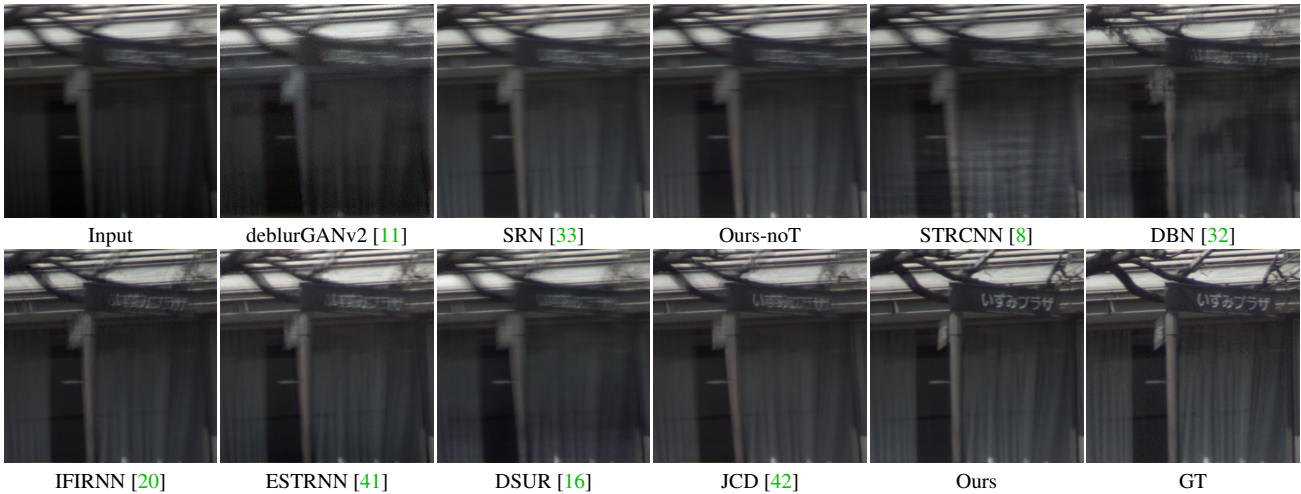

	\centering
	\footnotesize

    \begin{tabular}
    {   
        @{\hspace{0mm}}c@{\hspace{0.5mm}} 
        @{\hspace{0mm}}c@{\hspace{0.5mm}}
        @{\hspace{0mm}}c@{\hspace{0.5mm}} 
        @{\hspace{0mm}}c@{\hspace{0.5mm}} 
        @{\hspace{0mm}}c@{\hspace{0.5mm}} 
        @{\hspace{0mm}}c@{\hspace{0.5mm}} 
    }
        \cubrowy{1}{50}\\
    \end{tabular}
\caption{\textbf{Qualitative results.} Note that the scan direction is bottom-to-top ($\uparrow$).
}
\label{fig:result_large_1_50}
\end{figure*}

\section{Experiments}
\label{sec:results}

\subsection{Setup}

\paragraph{Dataset.} 
We use our contributed synthetic and real-world datasets for the following experiments. 
\begin{itemize}[leftmargin=*]
\vspace{-1mm}
\setlength\itemsep{0em}
\item \emph{Real-world dataset.} 
Using the proposed optic system, we collect the first RSGR dataset and the corresponding GS ground truth.
Our dataset consists of 79 video sequences 
captured under real scenes. Each sequence consists of 300 consecutive frames with $640\!\times\!640$ spatial resolution. The exposure time of the GS camera is 1ms, which is the same as the exposure time of the first scanline of the RSGR camera. Since the exposure time is short, the GS videos are sometimes slightly noisy. 
We split the dataset into a training set with 27 sequences and testing sets with 52 sequences.
The testing sets have two parts. The smaller one \textsc{Set-I} with 3 sequences share similar imaging conditions with the training set, while the larger one \textsc{Set-II} with 49 sequences have worse imaging conditions, where GS videos have noises.
We use \textsc{Set-I} as validation set.
Note that
prevalent datasets used for RS correction and GS deblur either are synthesized~\cite{Su-CVPR-2017-DBN,Liu-CVPR-2020-unrolling}, having significant gaps from real datasets or have no ground truth~\cite{Zhuang-ICCV-2017}, which are hard for developing and evaluating data-driven algorithms.
Unlike them, we capture real scenes with aligned ground truth (GT). We deliver them to the community to facilitate subsequent researches.

\item
\emph{Synthetic dataset.} 
We synthesize 25 video sequences for each of the GS, RS, and RSGR exposures.
Each video has 29 frames with $512\!\times\! 512$ resolution.
The scan directions are top-to-bottom ($\downarrow$).
The first scanlines of a corresponding GS, RS, and RSGR frame are exposed simultaneously and with the same duration. 
During synthesizing RSGR videos, we use a parameter $\xi$ to determine the ratio between readout time and the first scanline's exposure duration. 
We synthesize RSGR videos with 8 different $\xi$ for both training and testing.

\end{itemize}

\paragraph{Evaluation metrics.}
Our optic system offers the convenience of using captured GS videos as GT to assess different algorithms.
The Peak
Signal-to-Noise Ratio (PSNR) and the Structural Similarity Index Measure (SSIM)~\cite{Wang-TIP-2004-SSIM} are used as metrics.
We calculate them with a single video frame. Considering the distortion is spatial-varying, we also divide the video frame into parts for evaluation.

\subsection{External comparison}
\label{sec:exp_external}

\paragraph{Comparison with other algorithms.}
Since this problem is unexplored, we compare our algorithm with several closely related algorithms from four different categories: 
\textbf{1)} unsupervised GS image-based deblur algorithm that uses adversarial training to get rid of the paired training data: deblurGANv2~\cite{Kupyn-ICCV-2019-deblurGANv2};
\textbf{2)} supervised GS image-based deblur algorithm that integrates multi-scale reception fields: SRN~\cite{Tao-CVPR-2018-SRN};
\textbf{3)} supervised GS video-based deblur algorithms that leverages the temporal information by fusing neighboring frames or with an RNN: STRCNN~\cite{Hyun-ICCV-2017-STRCNN}, DBN~\cite{Su-CVPR-2017-DBN}, IFIRNN~\cite{Nah-CVPR-2019-IFIRNN}, and ESTRNN~\cite{Zhong-ECCV-2020};
and \textbf{4)} supervised RS correction or deblur methods that end-to-end estimate motion and compensate distortion: DSUR~\cite{Liu-CVPR-2020-unrolling} and JCD~\cite{Zhong-CVPR-2021}.

\tabref{compare-img-sota} reports the qualitative results.
We observe that all algorithms except for deblurGANv2 improve the original videos.
It demonstrates that supervised image-to-image translation algorithms can remove the RSGR distortion. But, as verified by Liu~\etal~\cite{Liu-CVPR-2020-unrolling}, they cannot correct the RS distortions since a rectified pixel in the virtual GS image might lie far away from its corresponding pixel in the input RS image.
The RS distortion correction requires accurately estimated motion.
Moreover, our algorithm achieves the \emph{best} performance.
It consistently outperforms others with a large margin on all evaluation metrics.
Especially, opposite to other algorithms that tend to improve only high-quality inputs captured without motion, textureless frames, \etc, our algorithm improves not only the high-quality inputs but also the low-quality counterparts
(see {{\emph{supplemental material}}}).
We suppose it is because the competing algorithms target RS or GS inputs with the same exposure duration for every pixel.
Differently, our RSGR video has gradually increased exposure duration along the scan direction, leading to spatially-varying blur and brightness distortion.
Our tailored algorithm incorporates a spatial-aware encoder and dual temporal information aggregators to achieve the best performance for correcting the RSGR distortion. 

\figref{result_large_1_50} presents qualitative results.
Similar to quantitative results, we find that our algorithm renders visually pleasing results that are better than other algorithms.
It corrects mixed spatial-varying blur and brightness \emph{without} introducing additional distortions.
Remarkably, thanks to our carefully selected loss function, our rendered videos are noise-free even though there are noises in the supervisory signal.
Unfortunately, other algorithms not for this problem have sub-optimal performance.
They fail to remove the distortions perfectly like us and even bring some artifacts, \eg, geometric deformations, color distortions, and noises.
For example, 
deblurGANv2 introduces noises due to adversarial training with noisy supervision.
STRCNN and DBN introduce color distortions.
IFIRNN and JCD yield geometric deformations owing to the large spatial-varying blur.

\begin{figure}[!t]
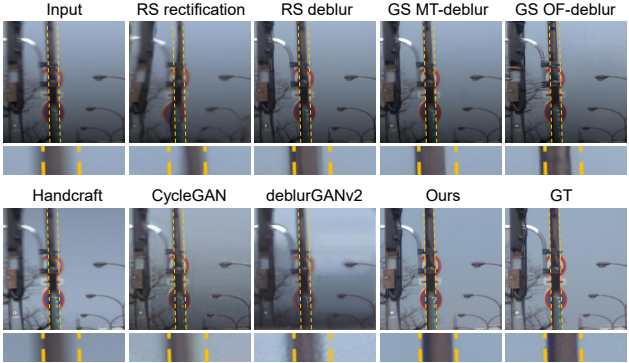

	\centering
	\footnotesize
    \mfigure{1}{exp/compare_knowledge7.pdf}
    \vspace{-3mm}
\caption{\textbf{Comparison with using different knowledge.} Note that the scan direction is bottom-to-top ($\uparrow$).}
\label{fig:compare_knowledge}
\end{figure}

\paragraph{Comparison with using other knowledge.}
We have illustrated that our tailored architecture is superior in RSGR correction.
But, it should be noted that besides the architecture, there also exists another factor accounting for this success: the \emph{paired} training data, which gives our model sufficient supervision to learn the \emph{task-specific} knowledge.
We seek to figure out if others can replace the knowledge. 
First, we compare RSGR correction using our end-to-end learned knowledge with that borrowed from other tasks (models with pre-trained weights) directly, including RS correction~\cite{Liu-CVPR-2020-unrolling}, RS deblur~\cite{Zhong-CVPR-2021}, GS motion deblur (MT-deblur)~\cite{Kupyn-ICCV-2019-deblurGANv2}, and GS out-of-focus deblur (OF-deblur)~\cite{Lee-CVPR-2021_IFAN}.
From the results in \figref{compare_knowledge}, we observe that although using pre-trained knowledge
directly relaxes the requirement for a paired dataset, they cannot deal with the spatial-varying brightness and blur.
Especially, the spatial-varying blurs lead to annoying geometric distortions.
The results confirm our required knowledge is different from other tasks.
Second, we use handcraft knowledge instead of learning-based knowledge to correct the spatial-varying brightness distortion. The result is frustrating.
Third, we train two unsupervised methods, \ie, CycleGAN~\cite{Zhu-ICCV-2017-CycleGAN} and deblurGANv2~\cite{Kupyn-ICCV-2019-deblurGANv2} without paired data.
The results verify our required knowledge cannot be replaced by unsupervised knowledge.
Accordingly, we build the optic system and paired data with incredible efforts.

\begin{figure}[!t]
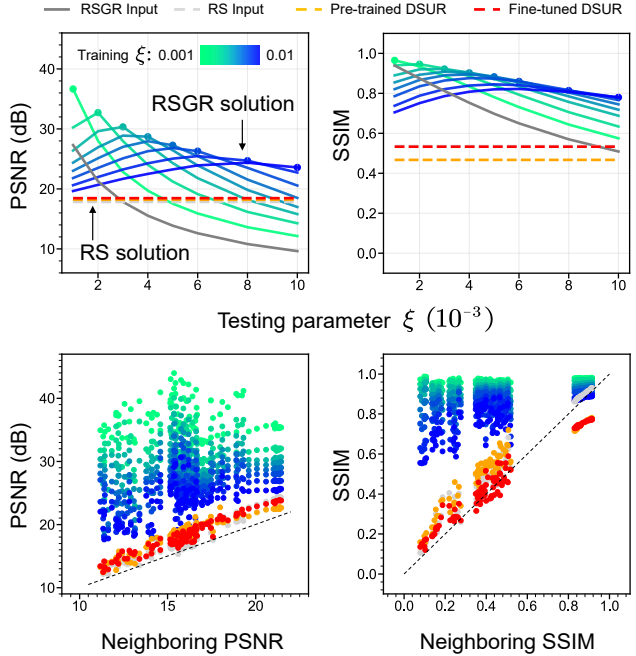

	\centering
	\footnotesize
    \mfigure{1}{exp/simulate_v6.pdf}
    \vspace{-3mm}
\caption{\textbf{Comparison with RS solution.}
We train and evaluate DSUR~\cite{Liu-CVPR-2020-unrolling} with synthesized RS videos as inputs and synthesized GS videos as supervision. Likewise, we train and evaluate our model with 8 different RSGR videos, synthesized with different $\xi$. We use neighboring metrics to represent the motion degree.}
\label{fig:compare_RS}
\vspace{-2mm}
\end{figure}

\paragraph{Comparison with RS solution.}
We argue that our solution is better than the RS solution.
{\textbf{1)} Formulation. With the global reset feature, we convert the RS \emph{rectification} problem into a \emph{deblur-like} one, which can be solved with image-to-image translation algorithms while the RS rectification cannot.}
\textbf{2)} Effectiveness. 
The results in \figref{compare_RS} show that our RSGR solution outperforms the RS solution.
Especially, our solution is good at dealing with significant motion (lower neighboring PSNR/SSIM).
We also find that our solution goes ahead of the RS solution with a large margin when $\xi$ is small, indicating a large exposure duration.
Although our performance decays with the increase of $\xi$, we are still superior to the RS solution. Considering it is not always to require a large ratio $\xi$ unless we need to capture high-speed videos, our solution is possible to replace RS solution in actual video capture.
\textbf{3)} Efficiency. 
In addition to accuracy, our solution without explicit motion estimation is also more efficient than existing RS solutions.
Classical two-frame-based RS correction methods often require a few minutes.
Zhuang~\etal \cite{Zhuang-ICCV-2017} processing a $640\!\times\!480$ resolution frame requires 400 seconds,
DSUR requires 0.43 seconds,
JCD requires 0.83 seconds,
but we only need 0.04 seconds\footnote{\torevise{We conduct all run-time comparisons of DL-based methods, \ie DSUR, JCD, and ours on the same machine with NVIDIA Tesla V100 GPU and Intel(R) CPU@3.80GHz. 
The run-time result of Zhuang~\etal \cite{Zhuang-ICCV-2017} borrowed from DSUR is based on an Intel Core i7-7700K CPU.}}.
Considering its \emph{effectiveness} and \emph{efficiency}, we believe our RSGR solution can replace RS solutions.

\subsection{Internal comparison}
\label{sec:exp_internal}

\paragraph{Architecture ablation.}
We conduct ablation experiments on our network architecture by implementing 9 different variants to explore the effectiveness of each module.
From the results in \tabref{ablation}, we have three main findings.
First, removing any of the components of our method will weaken the performance. 
The result verifies that all components are necessary.
Second, using spatial-aware modules (T6) has better performance than using only the temporal counterparts (S3). 
We argue that this is because the power of the temporal information aggregators relies on clean feature maps produced by spatial-aware encoders.
With the spatial-aware design, the aggregators would gather distorted feature maps and lose their effectiveness.
Third, we surprisingly find that using only one path of the long-term aggregator (T1) is worse than using full (\emph{ours}) and without using the long-term temporal aggregator (T2). We assume this is because our algorithm uses long-term and short-term temporal information together. When we gather the long-term information from one direction, the information imbalance in a video sequence will trouble the short-term aggregation.
Consequently, our long-term and short-term aggregators gather information from forward and backward directions and yield the best performance.

\begin{figure}[!t]
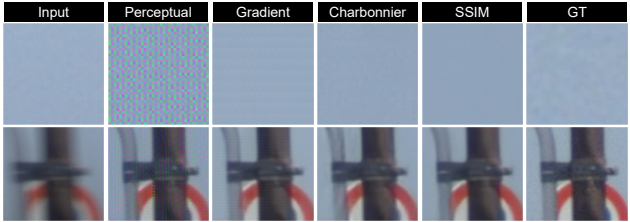

	\centering
    \mfigure{1}{exp/lossCompare_v4.pdf}
    \vspace{-5mm}
\caption{\textbf{Ablation experiments of different losses.}  Top patches have $37\!\times\!37$ resolution and bottom patches have $140\!\times\!129$ resolution. They come from the same frame. 
}
\label{fig:loss_compare}
\end{figure}

\paragraph{Loss ablation.}
We also verify the effectiveness of different loss functions in \figref{loss_compare}, including the perceptual loss~\cite{Johnson-ECCV-2016-Perceptual}, the gradient loss~\cite{Ma-2020-CVPR}, the Charbonnier loss~\cite{Charbonnier-ICIP-1994-Charbonnier} and the SSIM loss~\cite{Wang-TIP-2004-SSIM}.
We find that all losses other than the SSIM loss result in different artifacts. It is because there exist noises in the supervisory signal. 
This phenomenon also appears in the adversarial loss.
Besides, the perceptual loss has the best performance in structure restoration since it operates on the feature level.
Therefore, we combine the SSIM loss and the perceptual loss, leading to the best results.

\begin{table}[!tb]
    \centering
    \caption{\textbf{Ablation experiments of different architectures.}
    \footnotesize{Notations: S1 is our model without (\emph{w/o}) EE; S2 is \emph{w/o} SA; S3 is \emph{w/o} EE and SA; T1 is \emph{w/o} the backward path of the long-term aggregator; T2 is \emph{w/o} the long-term aggregator; T3 is replacing DCN of our model with Conv; T4 is combination of T2 and T3; T5 is T4 replacing CA with Conv; T6 is T5 \emph{w/o} short-term temporal information.}
    }
    \vspace{-2mm}
    \resizebox{\columnwidth}{!}
    {
    \begin{tabular}{lcccc}
        \toprule
        {Exp.} & 
        F & U & M & L \\
        \midrule
        \rowcolor{LightCyan}
        \emph{ours} & 
        32.72 / 0.92 &
        31.83 / 0.92 &
        33.01 / 0.92 & 
        34.65 / 0.92 \\ 
        \midrule
        S1 & 
        32.53 / 0.91 &
        31.28 / 0.90 & 
        32.45 / 0.91 &
        34.56 / 0.91 \\
        \rowcolor{LightCyan}
        S2 & 
        26.51 / 0.90 &
        28.58 / 0.90 &
        30.67 / 0.92 & 
        25.66 / 0.89 \\
        S3 & 
        25.09 / 0.89 & 
        25.39 / 0.88 & 
        29.56 / 0.91 & 
        24.70 / 0.89 
        \\
        \midrule
        \rowcolor{LightCyan}
        T1 & 
        31.76 / 0.90 & 
        30.60 / 0.89 & 
        32.15 / 0.90 & 
        34.10 / 0.91 \\
        T2 &  
        31.99 / 0.90 & 
        30.89 / 0.89 & 
        32.37 / 0.91 & 
        34.47 / 0.91 \\
        \rowcolor{LightCyan}
        T3 & 
        31.70 / 0.90 &  
        30.68 / 0.89 & 
        31.99 / 0.90 & 
        33.83 / 0.91 \\
        T4 &
        31.25 / 0.90 &
        29.91 / 0.89 &
        31.58 / 0.90 & 
        34.10 / 0.91 \\
        \rowcolor{LightCyan}
        T5 &
        29.31 / 0.89 & 
        28.12 / 0.88 & 
        29.75 / 0.89 & 
        32.48 / 0.90 \\
        T6 & 
        27.56 / 0.85 & 
        26.23 / 0.83 & 
        27.55 / 0.85 & 
        31.55 / 0.88 \\
        \midrule
        \rowcolor{LightCyan}
        \emph{input} & 
        18.95 / 0.75 &
        25.32 / 0.82 &
        21.56 / 0.81 & 
        16.36 / 0.63 
        \\
        \bottomrule
    \end{tabular}
    }
    \label{tab:ablation}
\end{table}

\subsection{Practicality evaluation} 
\label{sec:prac}

\begin{figure}[!tb]
    \centering
    \includegraphics[width=1.0\linewidth]{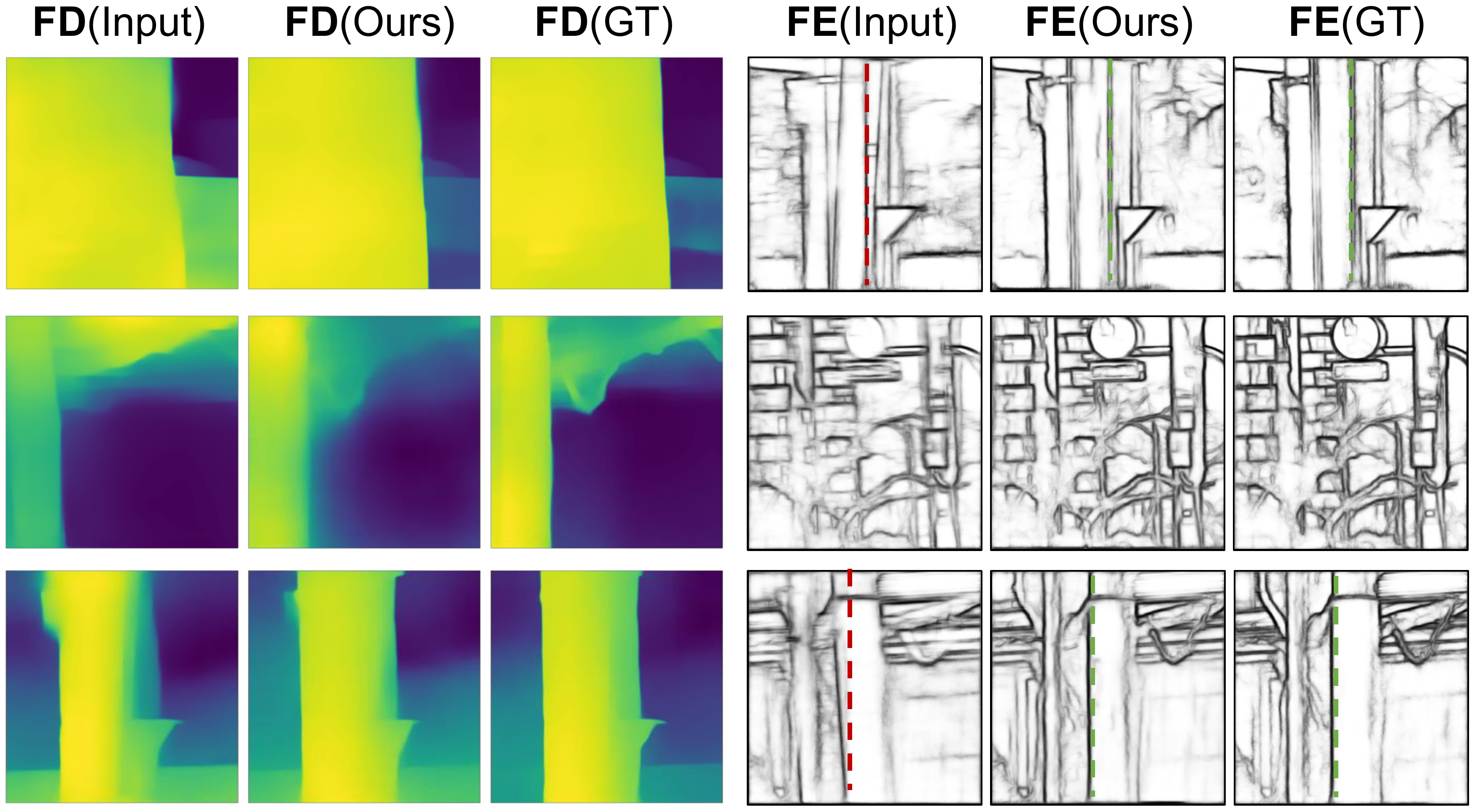}
    \vspace{-5mm}
    \caption{\textbf{Downstream applications.} 
    We apply two tasks with pre-trained models including \textbf{FD}: the single image depth estimation~\cite{Ranftl-TPAMI-2020-Depth} and \textbf{FE}:
    the edge detection~\cite{Liu--PAMI-2019-Edge} on the original RSGR, our corrected RSGR, and the GS video frames.
    }
    \label{fig:application}
\end{figure}

\begin{figure}[!tb]
    \centering
    \includegraphics[width=1.0\linewidth]{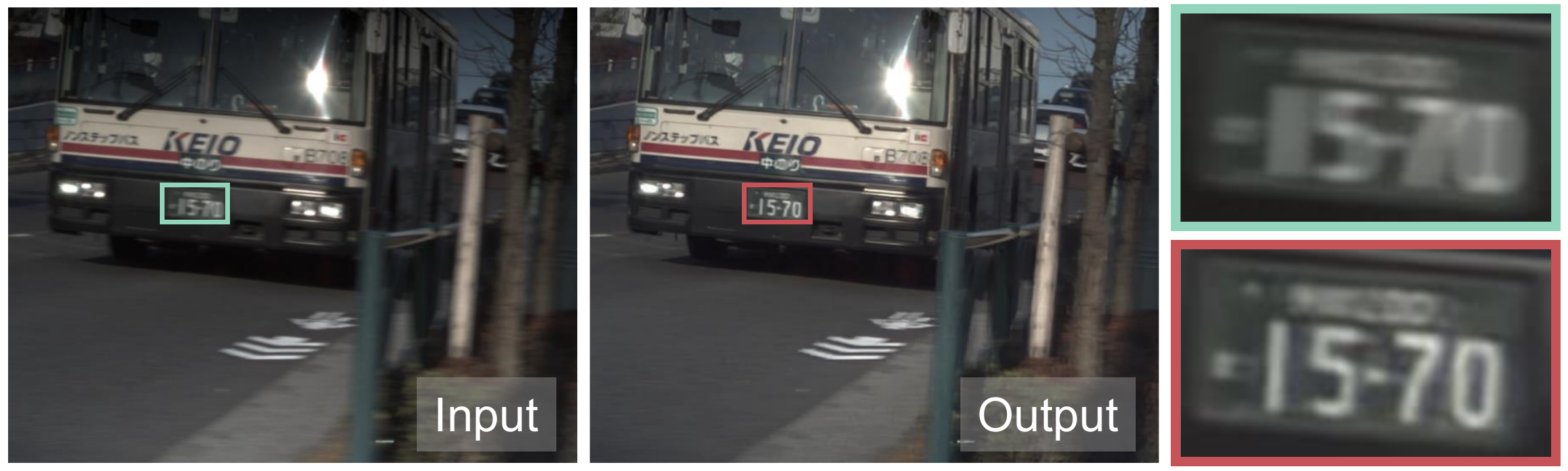}
    \vspace{-5mm}
    \caption{\textbf{Generalization evaluation.} We train our model on a certain RSGR camera and test it on another different RSGR camera, which has different exposure time, readout time, resolution, \etc. Note that the scan direction is top-to-bottom ($\downarrow$).
    }
    \label{fig:generation}
\end{figure}

Our RSGR solution works well on real applications.
Firstly, the rendered results are not only visually pleasing but also applicable to downstream tasks.
In \figref{application}, we perform single image depth estimation~\cite{Ranftl-TPAMI-2020-Depth} and edge detection~\cite{Liu--PAMI-2019-Edge}. The results reveal that although downstream tasks act worse on original RSGR videos captured undergoing motion, our produced virtual GS videos significantly improve them. The performance is even comparable to that of using GS videos.
Secondly, our algorithm has a favorable generalization ability.
The results in \figref{generation} prove that our algorithm trained on a specific RSGR camera can be applied directly to a different RSGR camera.
In addition, the results of \textsc{Set-II} (as \tabref{compare-img-sota} shows) can demonstrate the generalization ability of our algorithm across different imaging conditions. We still outperform other methods with a large margin when the testing set has different imaging conditions from the training set. 
Thirdly, as discussed in \secref{exp_external}, our solution is efficient. Without explicit motion estimation, it is $\times 10$ faster than the state-of-the-art RS solution DSUR~\cite{Liu-CVPR-2020-unrolling}.

\section{Conclusion}
\label{sec:conclusions}

This paper first attempts to restore clean GS videos from the RS sensor with the widely ignored global reset feature.
We build a new optic system and capture a new dataset with it. 
Based on the dataset, we developed a data-driven algorithm.
We experimentally demonstrate, with this feature, we can convert the RS rectification problem into a deblur-like one, getting rid of the non-trivial motion estimation steps.
The results also verify that our tailored algorithm achieves the best performance, potentially applying to real scenarios.
Compared to the RS solution, our solution is effective and efficient.
Considering it is easy to realize without changing the hardware, we believe our RSGR solution can replace the RS solution.

\paragraph{Limitations.}
Although we advocate RSGR over standard RS solution, it should be noted that RS videos do not suffer from any distortion in the completely static situation, yet RSGR still has intensity variations. Although our proposed algorithm can compensate it somehow, we suggest using our RSGR solution for dynamic scenarios. Relating to the point above, the last scanning lines of RSGR are more likely to be overexposed. We thus recommend adjusting the exposure time of RSGR properly. We also believe that the dynamic range might be improved by leveraging no-local exposure variations of RSGR. We leave it to our future work.

\vspace{-1mm}
{
\small
\paragraph{\small Acknowledgments.} This research is supported in part by the JSPS KAKENHI Grant Numbers 20H05951, 20H04215, the Key Project of Natural Science Research of Universities in Anhui (KJ2017A934), and the Value Exchange Engineering, a joint research project between Mercari, Inc. and RIISE. ZW also thanks the MEXT Scholarship.}

{\small
\bibliographystyle{ieee_fullname}
\bibliography{main}
}

\end{document}